\documentclass[10pt,twocolumn,letterpaper]{article}

\usepackage{wacv}
\usepackage{times}
\usepackage{epsfig}
\usepackage{graphicx}
\usepackage{epstopdf}
\usepackage{amsmath}
\usepackage{amssymb}
\usepackage{enumerate}
\usepackage{enumitem}
\usepackage{nopageno}
\usepackage{array}
\usepackage{amsthm,amsmath,amssymb}
\usepackage{mathrsfs}
\usepackage{multirow}
\usepackage{wrapfig}
\usepackage{float}
\usepackage[flushleft]{threeparttable}
\usepackage[ruled,lined]{algorithm2e}
\newcommand{\Comment}[1]{$\triangleright$\textit{#1}}

\def\wacvPaperID{589} 

\wacvfinalcopy 

\ifwacvfinal
\def\assignedStartPage{9876} 
\fi

\ifwacvfinal
\usepackage[breaklinks=true,bookmarks=false]{hyperref}
\else
\usepackage[pagebackref=true,breaklinks=true,colorlinks,bookmarks=false]{hyperref}
\fi

\ifwacvfinal
\else
\fi

\begin{document}

\title{Towards Resolving the Challenge of Long-tail Distribution \\in UAV Images for Object Detection}

\author{Weiping Yu\thanks{Equal contribution}, \hspace{0.25em}  Taojiannan Yang\footnotemark[1], \hspace{0.25em} Chen Chen\\
University of North Carolina at Charlotte\\
{\tt\small \{wyu4, tyang30, chen.chen\}@uncc.edu }
}
\maketitle

\begin{abstract}
Existing methods for object detection in UAV images ignored an important challenge -- imbalanced class distribution in UAV images -- which leads to poor performance on tail classes. We systematically investigate existing solutions to long-tail problems and unveil that re-balancing methods that are effective on natural image datasets cannot be trivially applied to UAV datasets. To this end, we rethink long-tailed object detection in UAV images and propose the Dual Sampler and Head detection Network (DSHNet), which is the first work that aims to resolve long-tail distribution in UAV images. The key components in DSHNet include Class-Biased Samplers (CBS) and Bilateral Box Heads (BBH), which are developed to cope with tail classes and head classes in a dual-path manner. Without bells and whistles, DSHNet significantly boosts the performance of tail classes on different detection frameworks. Moreover, DSHNet significantly outperforms base detectors and generic approaches for long-tail problems on VisDrone and UAVDT datasets. It achieves new state-of-the-art performance when combining with image cropping methods. Code is available at \url{https://github.com/we1pingyu/DSHNet}

\end{abstract}

\begin{figure}[t]
\begin{center}
 \includegraphics[width=0.9\linewidth]{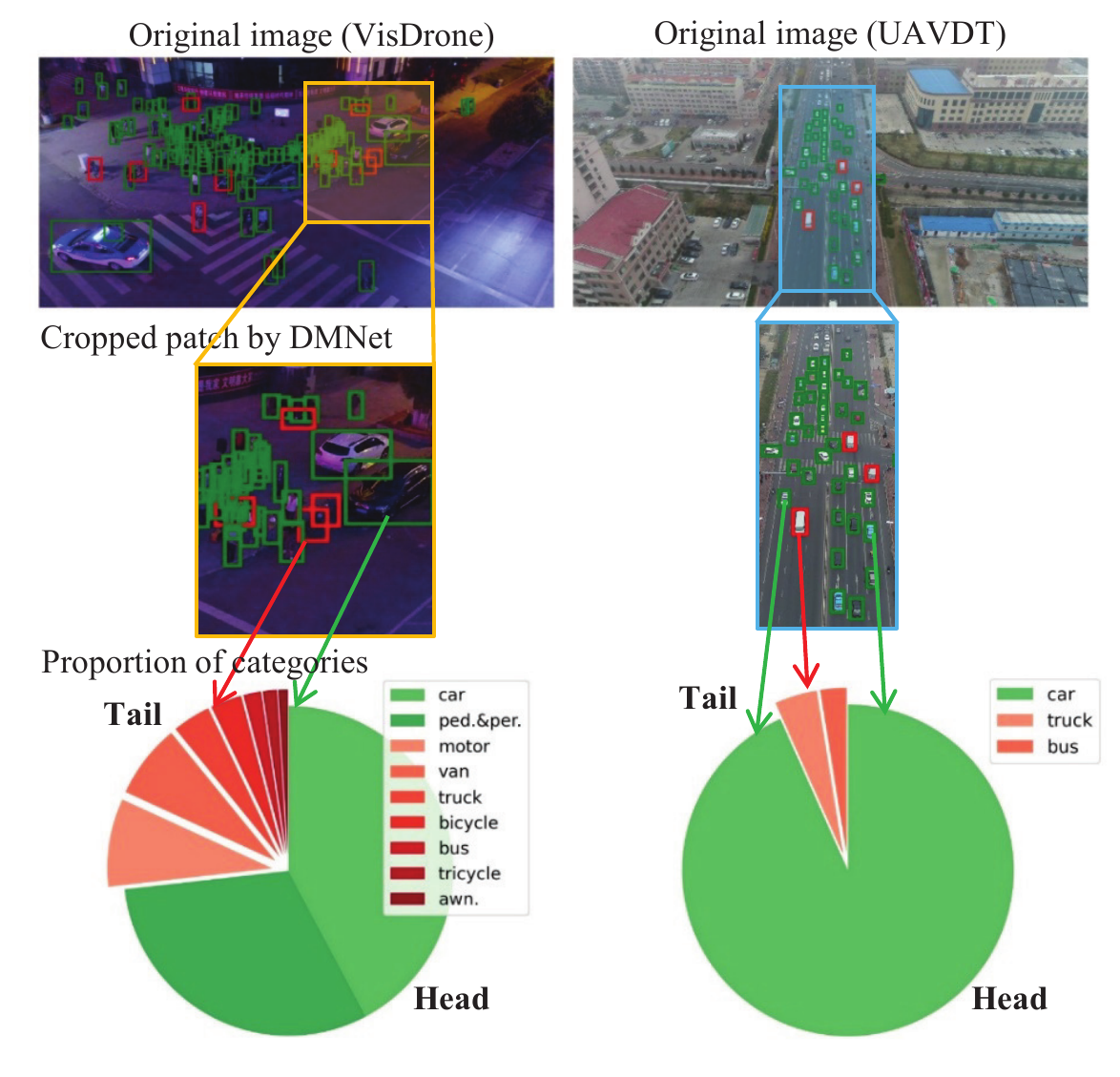}
\end{center}
\vspace{-0.5cm}
   \caption{UAV datasets (e.g., VisDrone~\cite{zhu2020vision} and UAVDT~\cite{du2018unmanned}) manifest the long-tail distribution phenomenon. The \textcolor{green}{green} boxes in these images denote the head-class objects and the \textcolor{red}{red} boxes denote the tail-class objects. The patches in the middle are generated by the DMNet~\cite{li2020density} cropping method. It can be observed that class imbalance exists in both original images and cropped patches. 
   In other words, cropping methods only solve the problem of spatial nonuniform distribution of targets, but does not take into account the issue of class imbalance. 
}
\vspace{-.2cm}
\label{fig:fig1}
\end{figure}

\begin{table*}
\footnotesize
\begin{center}
\begin{tabular}{|l|c|c|c|c|c|c|c|c|c|c|c|c|}
\hline
&\textcolor{blue}{ped}&\textcolor{blue}{person}&\textcolor{red}{bicycle}&\textcolor{blue}{car}&\textcolor{red}{van}&\textcolor{red}{truck}&\textcolor{red}{tricycle}&\textcolor{red}{awn.}&\textcolor{red}{bus}&\textcolor{red}{motor}\\
\hline\hline
Number of targets&79337&27059&10480&144867&24956&12875&4812&3246&5926&29647\\
Proportion (\%) &23.1&7.9&3.1&42.2&7.2&3.8&1.4&0.9 &0.7&8.6\\
\hline\hline
FRCNN (average precision) &21.4&15.6&6.7&51.7&29.5&19.0&13.1&7.7&31.4&20.7\\
\hline
\end{tabular}
\end{center}
\caption{Imbalanced distributions of objects/targets in VisDrone \cite{zhu2020vision} training set. Objects marked in {\color{red}{red}} color belong to the tail classes while those in {\color{blue}{blue}} belong to the head classes. \textbf{Note}: $ped$ (pedestrian) and $person$ are semantically the same, therefore we put $person$ in head classes as $ped$. In this table, we also show the average precision of each individual class based on Faster R-CNN~\cite{ren2015faster} (FRCNN for short). It is evident that there is a big performance gap between head classes and tail classes (e.g., \textcolor{blue}{car} vs. \textcolor{red}{awn.} $=$ 51.7 vs. 7.7). }
\label{table1}
\end{table*}

\section{Introduction}
\label{sec:introduction}
With the advance of unmanned aerial vehicles (UAVs), collecting high-quality images from the air has become convenient. Object detection plays a crucial role in many UAV applications, being a theme common to security and surveillance, infrastructure inspection and emergency response, among others. 
Although state-of-the-art deep learning-based object detectors (e.g., Faster R-CNN~\cite{ren2015faster}, Cascade R-CNN~\cite{cai2018cascade} and RetinaNet~\cite{lin2017focal}) can be directly applied to UAV datasets (e.g., VisDrone~\cite{zhu2020vision} and UAVDT~\cite{du2018unmanned}), previous studies~\cite{li2020density,yang2019clustered,hong2019patch} reveal that these detectors do not perform well on drone-captured scenes due to two main challenges: (1) targets appear very small in high-resolution UAV images; (2) targets have a nonuniform spatial distribution in images.  
To address these challenges, a great amount of effort has been witnessed on augmenting the input images, such as developing effective image cropping strategies~\cite{li2020density,yang2019clustered,hong2019patch,ozge2019power}.
For example, DMNet~\cite{li2020density} generates a density map for each image and utilizes it to crop the original image to patches based on the object density. 
The goal is to make objects evenly distributed in each cropped patch. The detection results are fused from the global image and cropped patches in the test phase. 

Although cropping methods alleviate the issue caused by uneven spatial distribution of the targets, \textit{they overlook another critical problem of UAV images}: object sample imbalance among categories. As shown in Table~\ref{table1}, a few classes in VisDrone~\cite{zhu2020vision} such as $car$, $ped.$ and $person$ account for more than 70\% of all the targets (these classes are known as head classes), while other classes such as $tricycle$ and $awn.$ have only a small number of samples (i.e., tail classes). This is referred to as the long-tail distribution. Based on the detection results of Faster R-CNN \cite{ren2015faster} in Table~\ref{table1}, we can easily notice the performance gap between head classes and tail classes (e.g., {car} vs. {awn.} $=$ 51.7 vs. 7.7 in terms of average precision). Fig.~\ref{fig:fig1} presents a few concrete examples to reveal the imbalanced class distribution in UAV images. The head-class objects (denoted by green boxes) dominate the scene. \textit{Even with the state-of-the-art cropping approach (DMNet \cite{li2020density}), the resulting patches still exhibit imbalanced class distribution. }



Tackling the long-tail distribution problem is important and has been a hot research topic in general object detection.
One common approach is to repeatedly sample targets of tail classes~\cite{han2005borderline} or discard some targets of head classes on purpose~\cite{drummond2003c4}. However, these straightforward methods suffer distortion of the original distribution which impairs the representation learning~\cite{zhou2020bbn,yang2020rethinking}. Therefore another line of research~\cite{zhou2020bbn,yang2020rethinking} aims to balance the class distribution by using different input distributions for training representation and classifier respectively in two phases (in the context of object detection, representation refers to feature extraction network and classifier refers to box head). These methods effectively reduce the performance drop caused by long-tail distribution on natural scene datasets (e.g., LVIS~\cite{gupta2019lvis}).


However, there are unique challenges that make the long-tail problem more difficult on UAV datasets than on general object detection datasets. Several state-of-the-art methods for long-tail visual recognition ~\cite{wang2020devil,yang2020rethinking} sample a balanced set of targets in a batch based on the assumption that they can use a relatively large batch-size. However, this assumption may not be warranted on UAV datasets because the images are high-resolution (e.g., VisDrone dataset has many images over the size of 2000$\times$1500). Due to the memory constraint, the typical batch-size is set to 1 or 2 for model training on these UAV datasets. Moreover, there are often hundreds of targets from head classes in one image (see 
the UAV images in Fig.~\ref{fig:fig1}). As a result, image-level repeat sampling~\cite{wang2020devil} is not an effective solution on UAV datasets.

In light of these challenges, we propose a novel Dual Sampler and Head Network (DSHNet) to address long-tail distribution in UAV datasets for object detection. DSHNet consists of two key components including Class-Biased Samplers (CBS) and Bilateral  Box Heads (BBH). Different from image-level re-sampling, CBS use the biased-sampling strategy to sample tail-class and head-class proposals separately with two samplers instead of a (default) single sampler in existing object detectors. BBH then separate the sampled tail-class and head-class proposals and process them with two box heads in the training phase to compute their losses respectively. In the test phase, each head of BBH only predicts results of the corresponding (tail or head) classes. 


Our main contributions can be summarized as follows:
\setlist{nolistsep}
\begin{itemize}[noitemsep,leftmargin=*]
\item We unveil the long-tail distribution problem in UAV images, which greatly hinders the performance of object detection. Previous works~\cite{yang2019clustered,li2020density,zhang2019dense} mainly focus on resolving the nonuniform spatial distribution by cropping the original images into small patches. However, the cropped images still exhibit long-tail distribution.

\item We analyze the unique challenges of object detection in UAV images and uncover that the solutions to long-tail distribution in natural images are not trivially applicable to UAV images. In light of this, we propose a novel Dual Sampler and Head Network (DSHNet) to handle head classes and tail classes separately.

\item DSHNet substantially outperforms the baseline models and generic long-tail solutions on various object detectors and network backbones. When coupled with image cropping methods (e.g., DMNet \cite{li2020density}), DSHNet further improves the detection performance, resulting in new state-of-the-art results on VisDrone and UAVDT benchmarks.



\end{itemize}

\section{Related work}
\label{sec:related}

\noindent \textbf{General object detection.} Deep learning-based object detection frameworks are divided into anchor-free and anchor-based ones. Anchor-free approaches~\cite{duan2019centernet,law2018cornernet} focus on detecting objects by locating and regressing key points. Anchor-based methods can be further grouped into two-stage~\cite{ren2015faster,cai2018cascade,he2017mask,dai2016r} and one-stage~\cite{lin2017focal,liu2016ssd,redmon2016you} detectors. The two-stage methods separate the training phase of detection into two steps: (1) using feature extraction network and anchor generator to produce candidate regions (i.e., region proposals); (2) utilizing box regression head to refine the results of step (1) and compute the loss. After the first step, the network usually adopts a sampler to sample some objects instead of training all of them to keep a balance between background and foreground proposals and to reduce computation. In one-stage methods, detectors directly regress the location and bounding box from anchors without candidate regions.

\noindent \textbf{Object detection in UAV images.} Compared with natural images, object detection in UAV images is more challenging. 
The performance of the generic object detectors is degenerated due to the spatial nonuniform distribution of targets and small target size. To tackle this issue, many approaches \cite{li2020density,yang2019clustered,hong2019patch,ozge2019power} generate a set of sub-images based on cropping methods. The general process of cropping-based methods is first using a proposal sub-net to analyze spatial information of objects and crop an image into small patches, and then training existing detection models with these patches. In the test phase, the final detection is obtained by fusing the detection results of local patches and global images with certain rules (e.g., non-maximum suppression (NMS)).

Another fundamental challenge lies in the imbalanced class distribution in UAV datasets, which leads to poor performance on tail classes as shown in Table~\ref{table1}. 
Only a few works ~\cite{chu2020feature,zhang2019dense} have touched upon this problem, yet they didn't address it from the perspective of solving long-tail distribution. For example, Zhang~\etal ~\cite{zhang2019dense} simply separate all the classes into two sub-categories and train two networks individually, which harms the generalization of representation and classifier due to discarding too many samples in training each network.
\begin{figure*}[t]
\begin{center}
 \includegraphics[width=0.85\linewidth]{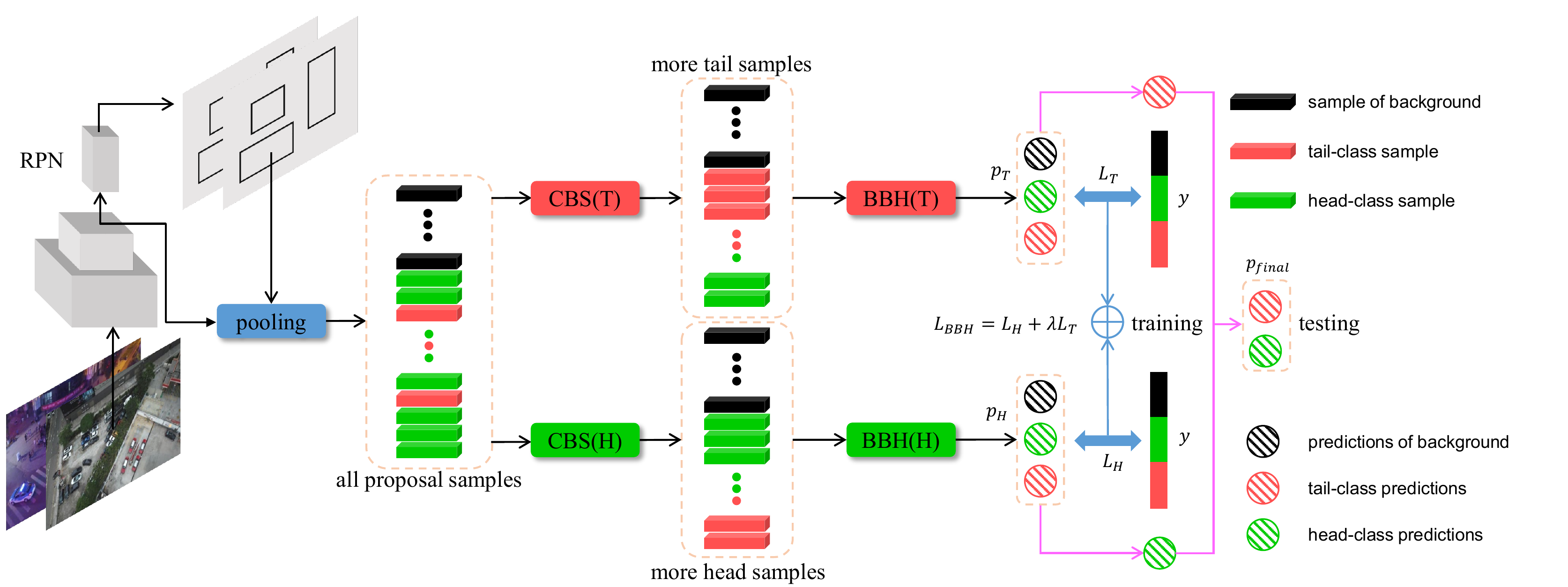}
\end{center}
\vspace{-0.2cm}
   \caption{The proposed DSHNet pipeline based on Faster R-CNN. RPN denotes region proposal network. CBS(*) denote tail(T)-biased and head(H)-biased samplers, and BBH(*) represent their corresponding box heads. The part after BBH shows that each box in BBH computes loss of all classes in the training phase, while only gives results of the corresponding (tail or head) classes in the inference phase.   
}
\label{fig:fig2}
\end{figure*}

\noindent \textbf{Long-tail object detection.} Most methods for long-tail object
detection~\cite{wang2020devil,li2020overcoming,wang2019classification} come from long-tail classification~\cite{yang2020rethinking,liu2019large,zhou2020bbn}, because the idea of dealing with the imbalanced class distribution is consistent. The following two approaches are considered to be the most effective ones:
\setlist{nolistsep}
\begin{itemize}[noitemsep,leftmargin=*]
\item {\bf Re-sampling}~\cite{buda2018systematic,byrd2019effect}: The main idea of re-sampling is to over-sample tail classes or under-sample head classes to balance the data distribution, thereby improving the chance of tail classes being trained. But sometimes, with re-sampling, duplicated samples of tail classes might lead to over-fitting, while discarding samples of head classes would impair the generalization ability of network.
\item {\bf Re-weighting}~\cite{cui2019class}: Re-weighting methods assign large weights for training samples of tail classes or hard instances in loss functions. However, re-weighting is not able to handle large-scale datasets since it can cause optimization difficulty~\cite{mikolov2013distributed}, leading to poor performance.
\end{itemize}

Beyond these methods, the authors of ~\cite{yang2020rethinking,zhou2020bbn} have shown that re-balanced input distribution improves classifier learning but harms feature learning. Therefore, many methods~\cite{zhou2020bbn,yang2020rethinking,wang2020devil} adopt the two-phase training paradigm: first train on the original data distribution normally; then fine-tune the classifier on a balanced data distribution with fixed representation. The current solutions to long-tail distribution are generally based on this paradigm.

But if we simply regard the long-tail issue in detection as the one in classification, then at least, we need to assume that in each batch, the numbers of targets of different classes are roughly the same in each image.
Unfortunately, this assumption is difficult to hold for object detection on UAV datasets since there is a huge gap in the numbers of targets of different classes as demonstrated in Fig.~\ref{fig:fig1}.

\section{Methodology}

\subsection{Long-tailed object detection in UAV images}
\label{sec:method:rethink}
We observe that existing re-sampling approaches for long-tailed object detection are based on the input of the model, namely image-level. To avoid over-fitting,  ~\cite{zhou2020bbn,wang2020devil,yang2020rethinking} propose to create a balanced input batch instead of simply repeating tail-class sampling or discarding head-class samples in each training iteration. These methods seem to alleviate the long-tail issue in general object detection, but their effectiveness depends on two requirements: (1) these methods require a relatively large batch-size so that there are enough targets of different categories to keep a balanced input batch; (2) they assume that each image contains a similar number of targets of different categories, otherwise the methods will degenerate to the simple repeat sampling or under-sampling which results in over-fitting or wasting training data. However, these two requirements are hard to meet in UAV object detection.

To be more specific, SimCal~\cite{wang2020devil}, a representative re-sampling approach, proposes to sample around 20\% classes in each batch (i.e., 16 out of 80 classes for MS COCO~\cite{lin2014microsoft}), so the batch-size is 16 where one image for each class. However, UAV images are usually in high resolution (e.g., a common resolution of 2000$\times$1500 for the VisDrone dataset). Therefore, only 1 or 2 images can be fit into one batch. Considering that the loss is 
computed by averaging over all sampled classes in each batch, a relatively large number of classes should be included in each batch. Even if the first requirement can be met by sufficient computing resources, the second requirement cannot hold in most UAV images. The long-tail issue in UAV images does not only occur among images (i.e., inter-image) but also within images (i.e., intra-image), because there are frequently too many objects belonging to the head classes in one image (see Fig.~\ref{fig:fig1} for an example). It is not because there are more head-class images than tail-class images as in general object detection scenarios. In this case, even if these methods have a balanced class distribution in terms of images, the objects of different classes are still highly imbalanced.


The Multi-Model Fusion (MMF) method in~\cite{zhang2019dense} is a specific solution to imbalanced class distribution in UAV images. It separates all the classes into two groups by the amount of targets in the dataset. Then two detectors are trained for the two groups of classes separately. In the test phase, MMF fuses the results from two detectors. MMF discards a lot of useful data in training each model, which 
impairs the model representation capability. For example, when the first model learns information about $car$, because the corresponding ground truth does not include $van$, the representation cannot learn the difference between the two  classes well.
So the performance only improves slightly over the base model (see Table~\ref{table2}). 

\subsection{Overview of DSHNet}

The proposed DSHNet is a plug-and-play method for all anchor-based one-stage or two-stage detectors. It has two key components: Class-Biased Samplers (CBS) and Bilateral Box Heads (BBH). The overall pipeline based on Faster R-CNN is depicted in Fig.~\ref{fig:fig2}. Different from the random sampler in Faster R-CNN, DSHNet has two biased samplers. CBS(H) samples head classes in priority while CBS(T) samples tail classes preferentially. After CBS, the two groups of biased samples are fed into BBH(H) and BBH(T) correspondingly. During training, BBH compute the loss over all classes which is the same as the box head in Faster R-CNN. While during inference, BBH only predict the results for the corresponding (head or tail) classes and the predictions are fused to obtain the final results. The feature extraction backbone and region proposal network (RPN) are the same as Faster R-CNN.


\subsection{Class-biased samplers}

To handle intra-image object imbalance in UAV images, we propose the Class-Biased Samplers (CBS) to perform re-sampling on the object-level when generating object proposals. The CBS module consists of two biased proposal samplers. The first one is CBS(T) which samples tail classes in priority. This means we first collect proposals which belong to the tail classes. If the number of proposals is insufficient after collecting all the tail-class proposals, CBS(T) continues to collect other classes. For example, in Faster R-CNN, the sampler will randomly sample 512 proposals, where 25\% of them are positive samples and 75\% of them are negative samples. CBS(T) will sample tail classes first. After that if the number of proposals is less than 128, the sampler will continue to sample targets of head classes. The second biased sampler is CBS(H), which works in the same scheme as CBS(T) by changing tail classes to head classes. Please refer to Alg.~\ref{algorithm1} for more details. It should be noted that the performance improvement of CBS is not due to the increased number of samples. As shown in the ablation study in Sec.\ref{sec:exp:ablation}, increasing the number of samples in the random sampler can not benefit the performance.
\begin{algorithm}[t]\footnotesize
\caption{Class-biased samplers}
\label{algorithm1}
\KwIn{samples of tail classes $S_t$, samples of head classes $S_h$, samples of background $S_b$, number of required samples $N_s$, positive fraction $\alpha$} 
\KwOut{Results of CBS(T) ${R_t}$, results of CBS(H) ${R_h}$}
$Random(set, num)$ means if $num<length(set)$ randomly return $num$ elements from $set$; else return all the $set$ \;
\Begin{
\Comment{initialization}

${R_t}\leftarrow\{\},{R_h}\leftarrow\{\}$\;
\Comment{compute numbers of required positive and negative samples}

$N_p=N_s \times \alpha, N_n=N_s-N_p$\;
\Comment{first add samples of background and tail classes to the results of CBS(T)}

${R_t} \leftarrow {R_t} \cup Random(S_b,N_n)$ \;
${R_t} \leftarrow {R_t} \cup Random(S_t,N_p)$ \;
\Comment{then add samples of head classes to the results of CBS(T) if needed}

\If{$length({S_t})<N_p$}{
${R_t} \leftarrow {R_t} \cup Random(S_h,N_p-length({S_t}))$\;
}
\Comment{repeat steps above for CBS(H) }

${R_h} \leftarrow {R_h} \cup Random(S_b,N_n)$ \;
${R_h} \leftarrow {R_h} \cup Random(S_h,N_p)$ \;
\If{$length({S_h})<N_p$}{
${R_h} \leftarrow {R_h} \cup Random(S_t,N_p-length({S_h}))$\;
}
}
\end{algorithm}



\subsection{Bilateral box heads}
Since the CBS module generates two groups of biased samples, we propose the Bilateral Box Heads (BBH) to process the tail-biased and head-biased proposals respectively. As demonstrated in \cite{zhou2020bbn, yang2020rethinking}, the backbone network (for feature extraction) prefers the original data distribution to learn well generalized representations. While the classifier tends to perform better on the biased classes in the data. By introducing BBH, we use two classifiers to cope with head classes and tail classes respectively. This allows each classifier to perform better on corresponding classes. Besides, CBS and BBH do not change the input distribution of the backbone network, which helps the backbone network to learn better generalized representations compared to existing image-level methods~\cite{zhou2020bbn,wang2020devil,wang2019classification}.




BBH can also re-weight the losses of head classes and tail classes. First, in each head of BBH, there are more samples of corresponding (head or tail) classes than those in the original box head, so the weight of corresponding classes is improved. Second, we can manually adjust the ratio between the losses of two heads. The final loss function of BBH is
\begin{equation}
\label{equation1}
L_{_{BBH}}={L}_{_H}(p_{_H},y)+\lambda L_{_T}(p_{_T},y),
\end{equation}
where ${L}_{_T}$ and ${L}_{_H}$ are the loss functions of BBH(T) and BBH(H), respectively.
$p_{_T}$ and $p_{_H}$ denote the predictions of BBH(T) and BBH(H) respectively, including box regression and class score. $y$ represents the label of bounding box and class. $\lambda$ is a balance coefficient. 


During training, both BBH(T) and BBH(H) compute losses on all classes since including other classes is beneficial to the generalization of the classifier. During inference, BBH(T) and BBH(H) only make predictions for its corresponding classes (see Fig.~\ref{fig:fig2}), and the two predictions are aggregated as the final results.

\subsection{Plug-and-play on base models} 
Fig.~\ref{fig:fig2} shows how to implement DSHNet on two-stage detectors. DSHNet can be easily applied to other main-stream detection pipelines including the one-stage and cascaded detectors without additional bells and whistles, as described in the following. 
\setlist{nolistsep}
\begin{itemize}[noitemsep,leftmargin=*]
\item {\bf One-stage detectors}: One-stage detectors directly perform regression and classification for every anchor in the image instead of generating ROIs (region of proposal, like Faster R-CNN~\cite{ren2015faster}). Take RetinaNet \cite{lin2017focal} for example, the final Retina head computes loss of all the targets. In DSHNet, we use CBS to assign label weights as 1 to sampled targets, so that other targets with label weights as 0 will not be considered in computing the loss.
\item {\bf Detectors with cascade architecture}: Cascade architecture means in the second stage of detection, networks have multiple (usually 3) box heads to process ROIs. Take Cascade R-CNN for example, in the second stage, there are 3 samplers for each box head. In DSHNet, we double all of them which means we have 6 samplers and 6 box heads (3 for tail classes and 3 for head classes). In the training phase, CBS and BBH work as same as them in Faster R-CNN. In the inference phase, class scores are computed by averaging over the 3 heads of BBH while box regressions are the results of the latest head of BBH of corresponding classes. 
\end{itemize}

\begin{table*}[t]
\begin{threeparttable}
\footnotesize
\begin{center}
\setlength\tabcolsep{3.5pt} 
\begin{tabular}{|l|c|c|c|c|c|c|c|c|c|c|c|c|c|c|c|}
\hline
Method&backbone&$AP$&$AP_{50}$&$AP_{75}$&ped.&person&bicycle&car&van&truck&tricycle&awn.&bus&motor\\
\hline\hline
\multicolumn{15}{|c|}{Comparison with base models}\\
\hline
RetinaNet+RS&R50&13.9&27.7&12,7&13.0&7.9&1.4&45.5&19.9&11.5&6.3&4.2&17.8&11.8\\
FRCNN+RS&R50&21.7&39.8&21.0&21.4&15.6&6.7&51.7&29.5&19.0&13.1&7.7&31.4&20.7\\
FRCNN+RS&R101&21.8&40.2&20.9&20.9&14.8&7.3&51.0&29.7&19.5&14.0&8.8&30.5&21.2\\
FRCNN+RS&X101&22.4&41.0&21.8&21.3&15.5&7.9&52.0&29.5&20.5&14.7&8.9&32.1&21.6\\
CRCNN+RS&R50&23.2&40.7&23.1&22.2&14.8&7.6&54.6&31.5&21.6&14.8&8.6&34.9&21.4\\
\hline
RetinaNet+DSHNet&R50&16.1&30.2&15.5&14.1&8.9&1.3&48.2&24.8&14.2&8.8&6.0&21.6&13.1\\
FRCNN+DSHNet&R50&24.6&44.4&24.1&22.5&16.5&10.1&52.8&32.6&22.1&17.5&8.8&39.5&23.7\\
FRCNN+DSHNet&R101&24.4&44.3&23.8&21.7&16.0&10.1&52.2&31.6&22.7&17.1&9.5&38.6&24.0\\
FRCNN+DSHNet&X101&25.8&\bf46.8&25.2&\bf23.3&\bf16.7&\bf11.4&53.7&33.1&23.8&\bf19.5&\bf11.1&40.0&\bf25.5\\
CRCNN+DSHNet&R50&\bf26.2&45.0&\bf26.3&23.2&16.1&11.2&\bf55.5&\bf33.5&\bf25.2&19.1&10.0&\bf43.0&25.1\\
\hline\hline
\multicolumn{15}{|c|}{Comparison with solutions to long-tail problems}\\
\hline
FRCNN+RS+MMF~\cite{zhang2019dense}&R50&22.6&41.7&21.6&21.6&15.3&9.6&51.5&28.5&20.4&15.9&7.5&33.7&21.6\\
FRCNN+SimCal~\cite{wang2020devil}&R50&20.0&35.8&19.6&18.7&13.8&5.7&51.0&28.4&16.4&13.6&5.9&27.0&19.4\\
FRCNN+RS+BGS~\cite{li2020overcoming}&R50&23.0&43.0&22.0&21.8&16.0&8.1&51.8&31.1&19.8&15.0&8.4&36.1&21.5\\
\hline
FRCNN+DSHNet&R50&\bf24.6&\bf44.4&\bf24.1&\bf22.5&\bf16.5&\bf10.1&\bf52.8&\bf32.6&\bf22.1&\bf17.5&\bf8.8&\bf39.5&\bf23.7\\
\hline\hline
\multicolumn{15}{|c|}{Based on the SOTA cropping method}\\
\hline
DMNet (FRCNN+RS)~\cite{li2020density}&R50&28.1$^{*}$&48.5&28.1&28.1&19.7&13.3&\bf57.3&36.1&24.8&20.1&12.0&42.9&26.4\\
\hline
DMNet~\cite{li2020density} cropping+DSHNet&R50&\bf30.3&\bf51.8&\bf30.9&\bf28.5&\bf20.4&\bf15.9&56.8&\bf37.9&\bf30.1&\bf22.6&\bf14.0&\bf47.1&\bf29.2\\
\hline
\end{tabular}
    \begin{tablenotes}
      \scriptsize
      \item {*} Note: we reproduce the DMNet~\cite{li2020density} result using authors' implementation and default settings since the original DMNet paper didn't report the class-wise AP. Our reproduced AP is 28.1, which is only 0.1 less than the result (28.2) in the original paper.
    \end{tablenotes}
\end{center}
\caption{The detection performance on VisDrone validation set. RS is short for random sampler.}
\label{table2}
\end{threeparttable}
\end{table*}

\section{Experiments}
\label{sec:experiment}
\subsection{Datasets}
To validate the effectiveness of DSHNet, we
conduct extensive experiments on two popular benchmarks for object detection in UAV images: VisDrone~\cite{zhu2020vision} and UAVDT~\cite{du2018unmanned}.
\begin{itemize}[leftmargin=*]
\item {\bf VisDrone}: The dataset consists of 10,209 images (6,471
for training, 548 for validation and 3,190 for testing) with rich
annotations on ten categories of objects. The image scale of the dataset is about 2,000 $\times$ 1,500 pixels. Since the evaluation server is closed now, we cannot test our method on the test set. Therefore, the validation set is used to evaluate our method, which is a setting adopted by previous methods as well.
\item {\bf UAVDT}: The UAVDT~\cite{du2018unmanned} dataset contains 23,258 images for training and 15,069 images for testing. The
resolution of the image is about 1,080 $\times$ 540 pixels. The
dataset is acquired with an UAV platform at a number of
locations in urban areas. The categories of the annotated
objects are $car$, $bus$, and $truck$.
\end{itemize}

\subsection{Implementation details}
\label{sec:exp:imple}
We implement DSHNet based on the MMdetection~\cite{mmdetection} toolbox. Faster R-CNN (FRCNN) ~\cite{ren2015faster}, RetinaNet (Retina)~\cite{lin2017focal} and Cascade R-CNN (CRCNN)~\cite{cai2018cascade} with Feature Pyramid Network (FPN)~\cite{lin2017feature} which are representatives of two-stage detectors, one-stage detectors and detectors with cascade architecture, are adopted as the baseline detection networks. The parameters of CBS are basically the same as the original random sampler (RS, randomly sample the corresponding number of positives and negatives) in the implementation of Faster R-CNN in MMdetection: the total number of samples is 512, and 25\% of the samples are positive and 75\% are negative. In VisDrone~\cite{zhu2020vision}, $car$, $ped.$ (pedestrian) and $people$ are considered as head classes (noted that the boundary between $ped.$ and $people$ in VisDrone is ambiguous, so we group them together), while other object categories are regraded as tail classes. In UAVDT~\cite{du2018unmanned}, $car$ is the head class and $truck$ and $bus$ are the tail classes. The BBH is the same as the original box head in Faster R-CNN which has 2 shared fully connected layers and 1 fully connected layer for predicting class scores and box regressions, respectively.

\noindent \textbf{Training phase.} The input size of the detector is
1,000 $\times$ 600 pixels on both VisDrone~\cite{zhu2020vision} and UAVDT~\cite{du2018unmanned} datasets. The batch-size is set to 2 (i.e., 2 images) on  a single NVIDIA 1080Ti GPU with 11GB memory. On VisDrone, we set the base learning rate to 0.002 and the training epochs to 18. After the 8th, 12nd and 16th epoch, the learning rate decreases by a factor of 10. On UAVDT, we set the training epochs to 4 and fix the learning rate at 0.002.

\noindent \textbf{Test phase.} The input size of detector is the same as that in the training phase. The maximum detection number is set to 500 by following the settings of VisDrone~\cite{zhu2020vision}. Following the evaluation protocol on MS COCO~\cite{lin2014microsoft}, we use $AP$, $AP_{50}$, and $AP_{75}$ as metrics to measure the precision. $AP_{50}$ and $AP_{75}$ are computed at the single IoU threshold 0.5 and 0.75 over all categories. In addition, we report the average precision of each object category.

\subsection{Experimental results}
\label{sec:exp:comparison}
\noindent \textbf{Comparison with base models.}
We implement 3 base models: Faster R-CNN (FRCNN)~\cite{ren2015faster}, RetinaNet (Retina)~\cite{lin2017focal} and Cascade R-CNN (CRCNN)~\cite{cai2018cascade} according to the same settings of DSHNet. For the backbone network, we choose ResNet50 (R50), ResNet101 (R101)~\cite{he2016deep} and ResNeXt101 (X101)~\cite{xie2017aggregated}, and use the default parameters (including the feature pyramid network (FPN)~\cite{lin2017feature}) in MMdetection~\cite{mmdetection}.

Table~\ref{table2} reports the overall $AP$ and $AP$s of all the 10 classes of VisDrone. DSHNet achieves consistent improvements on all the base models. On the most representative two-stage detector, Faster R-CNN, we conduct experiments using 3 different backbone networks, and the performance increases the most on ResNeXt101 (22.4 vs. 25.8). On the state-of-the-art Cascade R-CNN (CRCNN) model, our method also gains a considerable 3.0 overall AP (23.2 vs. 26.2). Moreover, in almost all the cases, AP of each class is improved compared with the counterpart in the base models. This demonstrates that both head and tail classes can benefit from DSHNet. For tail classes in particular, the detection results are improved significantly. For example, our DSHNet boosts the $AP$s of tail classes $tricycle$ and $bus$ by about 29\% (14.8 vs. 19.1) and 23.2\% (34.9 vs. 43.0), respectively, based on the CRCNN (R50). 
\begin{figure*}[h!]
\begin{center}
 \includegraphics[width=0.99\linewidth]{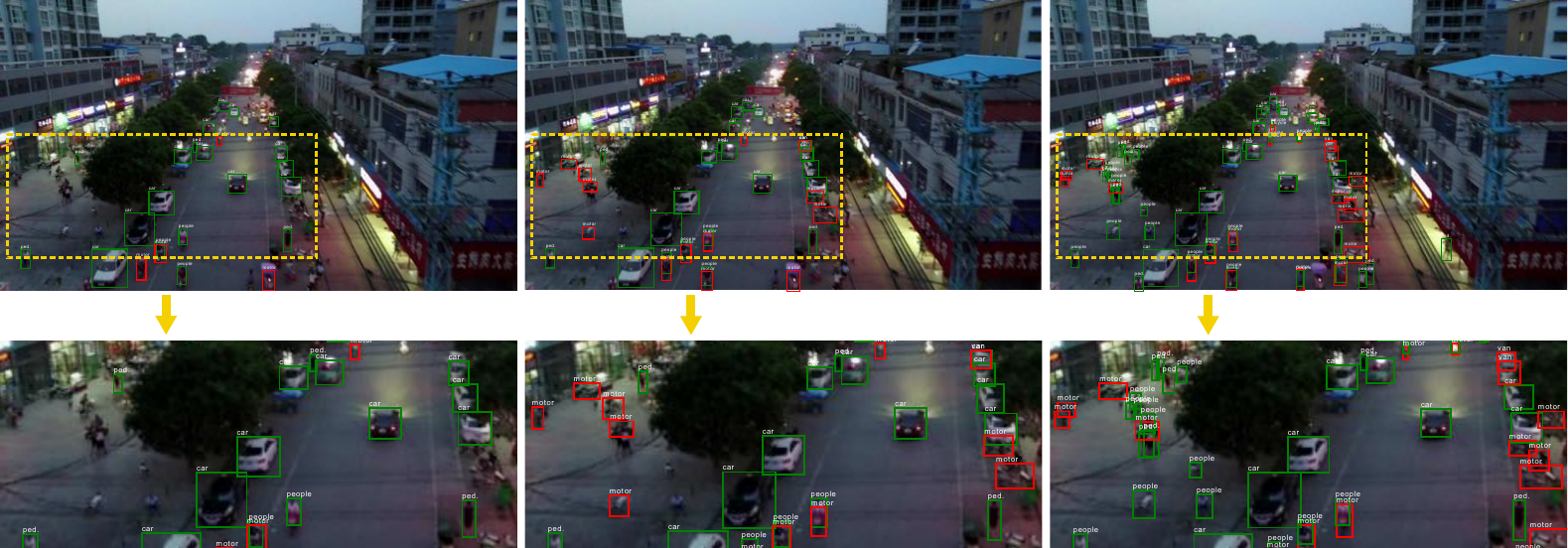}
\end{center}
\vspace{-0.1cm}
   \caption{Visualization of detection results on VisDrone. From the first to third columns are base model (Faster R-CNN (ResNet50)), DSHNet on the base model, and ground truth (\textcolor{red}{red} boxes for tail classes and \textcolor{green}{green} boxes for head classes). The detection precision of tail-class $motor$ is greatly improved by DSHNet. (Best viewed on screen with zoom. More examples in \textbf{Appendix}.)
}
\label{fig:vis}
\end{figure*}
We also present a visual comparison of the detection results on VisDrone dataset (Fig.~\ref{fig:vis}). In this example, the detection precision of $motor$, one of the tail classes, is obviously improved by DSHNet as compared to the base model.

\begin{table}
\footnotesize
\begin{center}
\setlength\tabcolsep{2.8pt} 
\begin{tabular}{|l|c|c|c|c|c|c|c|c|}
\hline
Method&backbone&$AP$&$AP_{50}$&$AP_{75}$&car&truck&bus\\
\hline\hline
FRCNN+RS&R50&11.0&19.9&11.2&28.3&1.4&3.4\\
FRCNN+RS&R101&12.5&22.3&12.9&27.9&1.9&7.8\\
Retina+RS&R50&14.5&28.5&12.9&30.0&3.7&9.7\\
\hline\hline
FRCNN+DSHNet&R50&16.0&28.5&16.3&29.4&3.9&14.7\\
FRCNN+DSHNet&R101&17.0&30.0&18.1&29.8&3.4&\bf17.9\\
Retina+DSHNet&R50&\bf17.8&\bf30.4&\bf19.7&\bf32.1&\bf4.2&17.0\\
\hline
\end{tabular}
\end{center}
\caption{The detection performance on UAVDT validation set.}
\label{tab:uavdt}
\end{table}

On UAVDT, we conduct experiments based on RetinaNet~\cite{lin2017focal} and Faster R-CNN~\cite{ren2015faster}. Similar to the results in VisDrone, DSHNet achieves remarkable improvements over baseline models, and the precision of both head classes and tail classes are largely improved. Specifically, the tail class $bus$ is dramatically improved from 3.4 to 14.7 on FRCNN-R50 as shown in Table \ref{tab:uavdt}. This clearly validates the effectiveness of our proposed DSHNet.

\begin{table*}
\footnotesize
\begin{center}
\setlength\tabcolsep{3.5pt} 
\begin{tabular}{|l|c|c|c|c|c|c|c|c|c|c|c|c|c|c|c|}
\hline
Method&backbone&$AP$&$AP_{50}$&$AP_{75}$&ped.&person&bicycle&car&van&truck&tricycle&awn.&bus&motor\\
\hline\hline
FRCNN+RS&R50&21.7&39.8&21.0&21.4&15.6&6.7&51.7&29.5&19.0&13.1&7.7&31.4&20.7\\
FRCNN+RS-DBL&R50&21.6&40.1&20.3&20.9&15.4&6.6&52.1&27.4&18.9&13.5&8.0&32.2&21.5\\
FRCNN+RS-DBL+BBH&R50&22.1&41.1&21.2&21.1&15.0&7.7&51.6&29.1&18.9&15.2&8.3&33.1&20.9\\
FRCNN+CES+BBH&R50&24.1&43.7&23.7&21.9&15.8&9.4&52.1&30.8&22.0&\bf17.7&\bf9.7&39.5&22.5\\
FRCNN+CBS+BBH&R50&\bf24.6&\bf44.4&\bf24.1&\bf22.5&\bf16.5&\bf10.1&\bf52.8&\bf32.6&\bf22.1&17.5&8.8&\bf39.5&\bf23.7\\
\hline
\end{tabular}
\end{center}
\vspace{-0.2cm}
\caption{The ablation study of CBS on VisDrone dataset.}
\label{tab:ablation:CBS}
\end{table*}

\begin{table*}
\footnotesize
\begin{center}
\setlength\tabcolsep{3.5pt} 
\begin{tabular}{|l|c|c|c|c|c|c|c|c|c|c|c|c|c|c|c|}
\hline
Method&backbone&$AP$&$AP_{50}$&$AP_{75}$&ped.&person&bicycle&car&van&truck&tricycle&awn.&bus&motor\\
\hline\hline
FRCNN+RS&R50&21.7&39.8&21.0&21.4&15.6&6.7&51.7&29.5&19.0&13.1&7.7&31.4&20.7\\
FRCNN+CBS&R50&21.7&40.2&20.8&20.5&14.8&6.9&51.6&29.0&19.1&13.9&8.3&31.6&20.7\\
FRCNN+CBS+BBH-ALL&R50&22.6&40.0&22.7&16.2&14.5&\bf10.1&47.8&32.0&19.6&17.1&\bf8.8&37.2&23.4\\
FRCNN+CBS+BBH&R50&\bf24.6&\bf44.4&\bf24.1&\bf22.5&\bf16.5&\bf10.1&\bf52.8&\bf32.6&\bf22.1&\bf17.5&\bf8.8&\bf39.5&\bf23.7\\
\hline
\end{tabular}
\end{center}
\vspace{-0.2cm}
\caption{The ablation study of BBH on VisDrone dataset.}
\label{tab:ablation:BBH}
\end{table*}

\noindent \textbf{Comparison with existing solutions tackling long-tail problems.}
We compare our DSHNet with state-of-the-art methods for long-tail visual recognition. In MMF~\cite{zhang2019dense} and BGS~\cite{li2020overcoming}, tail and head classes are defined according to the same rules as DSHNet (see Sec.~\ref{sec:exp:imple} for detail). In SimCal~\cite{wang2020devil}, we choose class-agnostic method which has better performance on overall AP. We assign 2 classes per batch for training in order to keep the sample ratio unchanged (2:10 on VisDrone~\cite{zhu2020vision} vs. 16:80 on MS COCO~\cite{lin2014microsoft}). 

We first compare our method with MMF~\cite{zhang2019dense}, which also addresses the imbalanced class distribution in VisDrone dataset. The main idea of MMF is to divide the 10 classes to 2 sub-categories and train and test the model separately for those two sub-categories.The feature network of each model in MMF only gets a part of all the classes as input, instead of getting the original distributed data like DSHNet, which harms the generalization of representation. Therefore, this method only slightly improves the base model from the results in Table~\ref{table2} (21.7 vs. 22.6). And the overall AP is 2 points worse than our DSHNet (22.6 vs. 24.6).

As discussed in Sec.~\ref{sec:related}, most existing methods tackling the long-tail problems on natural datasets do not work well on UAV datasets. Here, we choose two recent methods representing the re-sampling and re-weighting strategies, respectively. SimCal~\cite{wang2020devil} is improved from image-level re-sampling which also uses class-aware sampler. However, SimCal~\cite{wang2020devil} needs to sample enough categories and collect similar numbers of different targets to ensure a balanced distribution in a batch. SimCal~\cite{wang2020devil} is not trivially applicable on UAV datasets, and the performance is even inferior to that of the base model (20.0 vs. 21.7).

BGS~\cite{li2020overcoming} is the representative re-weighting method which introduces the balanced group softmax to group corresponding classes together when computing loss. BGS is currently the best re-weighting solution to long-tail issue. However, BGS only achieve re-weighting in loss function while does not change the distribution of box head, which proved to effectively improve classifier learning~\cite{wang2020devil,yang2020rethinking,zhou2020bbn}. Therefore, the imbalanced distribution of input cannot be re-balanced as good as DSHNet. Despite its better performance than the base model, BGS is still worse than DSHNet (23.0 vs. 24.6).

\noindent \textbf{How DSHNet performs with image patches?}
As shown in Fig.~\ref{fig:fig1}, image cropping methods do not solve the long-tail distribution problem. Since image cropping can be considered as a data pre-processing procedure, our DSHNet can be equally applied to image patches, which could further improve the performance of cropping-based approaches. To evaluate DSHNet on image patches, we employ the state-of-the-art DMNet~\cite{li2020density} to generate cropped patches and train DSHNet on them. The results in Table~\ref{table2} (last two rows) show that DSHNet also works on image patches and boosts the overall AP by 2.2 points (30.3 vs. 28.1).

\subsection{Ablation study}
\label{sec:exp:ablation}
To validate the contributions of CBS and BBH to the improvement of detection performance, we carry out ablation experiments on VisDrone~\cite{zhu2020vision} dataset with Faster R-CNN~\cite{ren2015faster} and ResNet50~\cite{he2016deep}.

\noindent \textbf{Effect of CBS.}
In order to verify that the performance of the detector is not simply related to the number of samples generated by the sampler, we conduct two experiments. First, we double the number of samples for the random sampler in Faster R-CNN~\cite{ren2015faster} (denoted by FRCNN+RS-DBL) in Table~\ref{tab:ablation:CBS}. The performance is on par with the base model (21.6 vs. 21.7), indicating that simply increasing the number of samples will not improve the performance when the sample size is adequate. Second, we remove the CBS module and use the random sampler to obtain two groups of samples as input to BBH (denoted by FRCNN+RS-DBL+BBH in Table~\ref{tab:ablation:CBS}). The performance is only slightly better than the base model (22.1 vs. 21.7), which verifies the crucial role of CBS in proposal re-sampling.

Another question we would like to answer is ``which sampler is better, class-biased or class-exclusive?". Class-exclusive sampler (CES) means that we only sample assigned classes without other classes. More concretely, after CBS(T) samples all the tail-class samples, if the number of samples does not meet the requirement, it will continue to sample head-class proposals, but CES will not.
The performance of CES is lower than CBS (24.1 vs. 24.6) which shows that if not affecting the main classes, adding other classes is beneficial to the generalization of the classifiers. 

\noindent \textbf{Effect of BBH.} To show that BBH can enhance the discriminative power of the classifiers, we remove BBH and simply integrate the results of CBS as input to the original single box head (FRCNN+CBS in Table~\ref{tab:ablation:BBH}). The performance of the network without BBH drops considerably compared to that of the complete DSHNet (21.7 vs. 24.6).

In the test phase, BBH(T) or BBH(H) only makes predictions for tail or head classes before the results aggregation. To validate the advantage of this prediction strategy in BBH, we establish a comparison method by allowing both BBH(T) and BBH(H) to generate predictions for all classes (denoted by FRCNN+CBS+BBH-ALL in Table~\ref{tab:ablation:BBH}) and fusing the predictions by NMS.
The separate prediction strategy improves the AP by 2 (22.6 vs. 24.6). It is because each head of BBH is trained with class-biased samples and by only considering the class predictions that they are good for (i.e., tail-class predictions from BBH(T) and head-class predictions from BBH(H)) would achieve better performance.


\setlength{\columnsep}{10pt}
\begin{wrapfigure}{r}{0.2\textwidth}
\vspace{-0.75cm}
\begin{center}
 \includegraphics[width=0.2\textwidth]{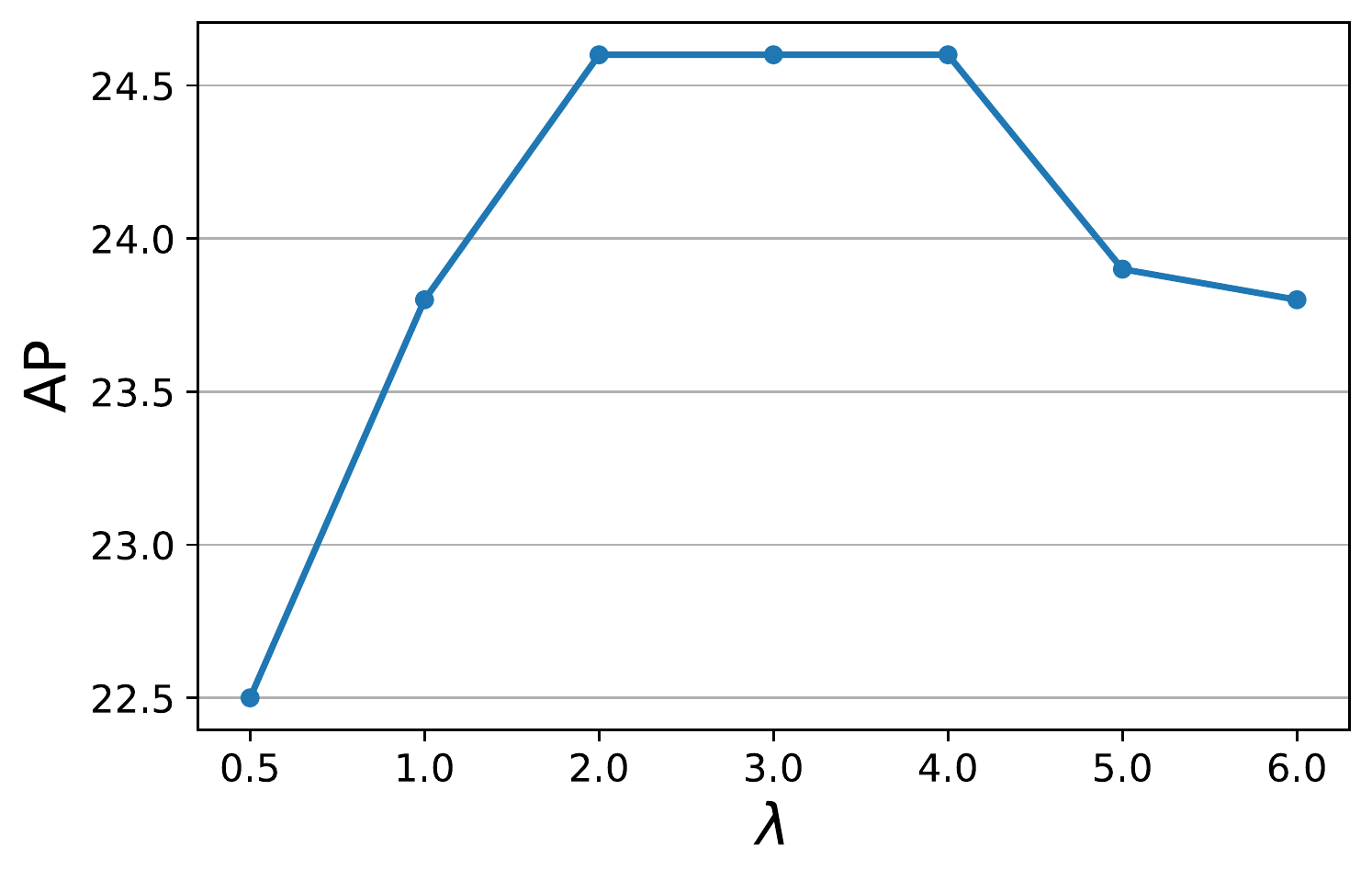}
\end{center}
\vspace{-0.25cm}
   \caption{\small{The AP over
different settings of $\lambda$ in the loss of BBH.}
}
\vspace{-0.25cm}
\label{fig:lambda}
\end{wrapfigure}

\noindent \textbf{Effect of $\lambda$.}
The coefficient $\lambda$ in the loss function of BBH (see Eq.~\ref{equation1}) can be optimized for performance. As shown in Fig.~\ref{fig:lambda}, when $\lambda<1$, the performance is relatively poor because the tail classes have less weight in the training. Between 2.0 and 4.0, the performance remains at the highest AP (24.6). Then the performance declines when $\lambda>4.0$. In the experiments, we set $\lambda=2.0$.

\noindent \textbf{Inference speed.}
We report the inference speeds of base models and DSHNet based on the implementations of MMdetection~\cite{mmdetection} on one NVIDIA 1080Ti GPU (see Table~\ref{tab:time}). DSHNet is only slightly slower than base models.
\vspace{-0.4cm}
\begin{table}[h!]
\footnotesize
\begin{center}
\setlength\tabcolsep{3.5pt} 
\begin{tabular}{|l|c|c|c|c|}
\hline
\multirow{2}*{Model}&\multicolumn{2}{c|}{base}&\multicolumn{2}{c|}{DSHNet}\\
\cline{2-5}
~&fps&$AP$&fps&$AP$\\
\hline\hline
RetinaNet~\cite{lin2017focal}&22.5&13.9&16.4&16.1\\
Faster R-CNN~\cite{ren2015faster}&20.1&21.7&15.7&24.6\\
Cascade R-CNN~\cite{cai2018cascade}&14.9&23.2&10.8&26.2\\
\hline
\end{tabular}
\end{center}
\vspace{-0.2cm}
\caption{The inference speed (frames per second) and $AP$ of base models and DSHNet with ResNet-50.}
\label{tab:time}
\end{table}

\vspace{-0.2cm}
\section{Conclusion}
\label{sec:conclusion}
We proposed a novel Dual Sampler and Head detection Network (DSHNet) to solve the long-tail distribution problem in UAV datasets. The Class-Biased Samplers (CBS) are introduced to perform biased sampling on object proposals for tail and head classes respectively.
The Bilateral Box Heads (BBH) use two classifiers to process the tail-biased and head-biased proposals separately.
Moreover, BBH achieve loss re-weighting by computing the loss for head and tail classes respectively. Experiments on two UAV benchmarks show that our method significantly improves the base models and achieves new state-of-the-art results. 

{\small
\bibliographystyle{ieee_fullname}
\bibliography{egbib}
}

\clearpage
\onecolumn
\appendix
{\centering{\section*{Appendix}}}
\renewcommand{\thefigure}{A\arabic{figure}}
\setcounter{figure}{0}

\begin{figure*}[h!]
\begin{center}
 \includegraphics[width=0.98\linewidth]{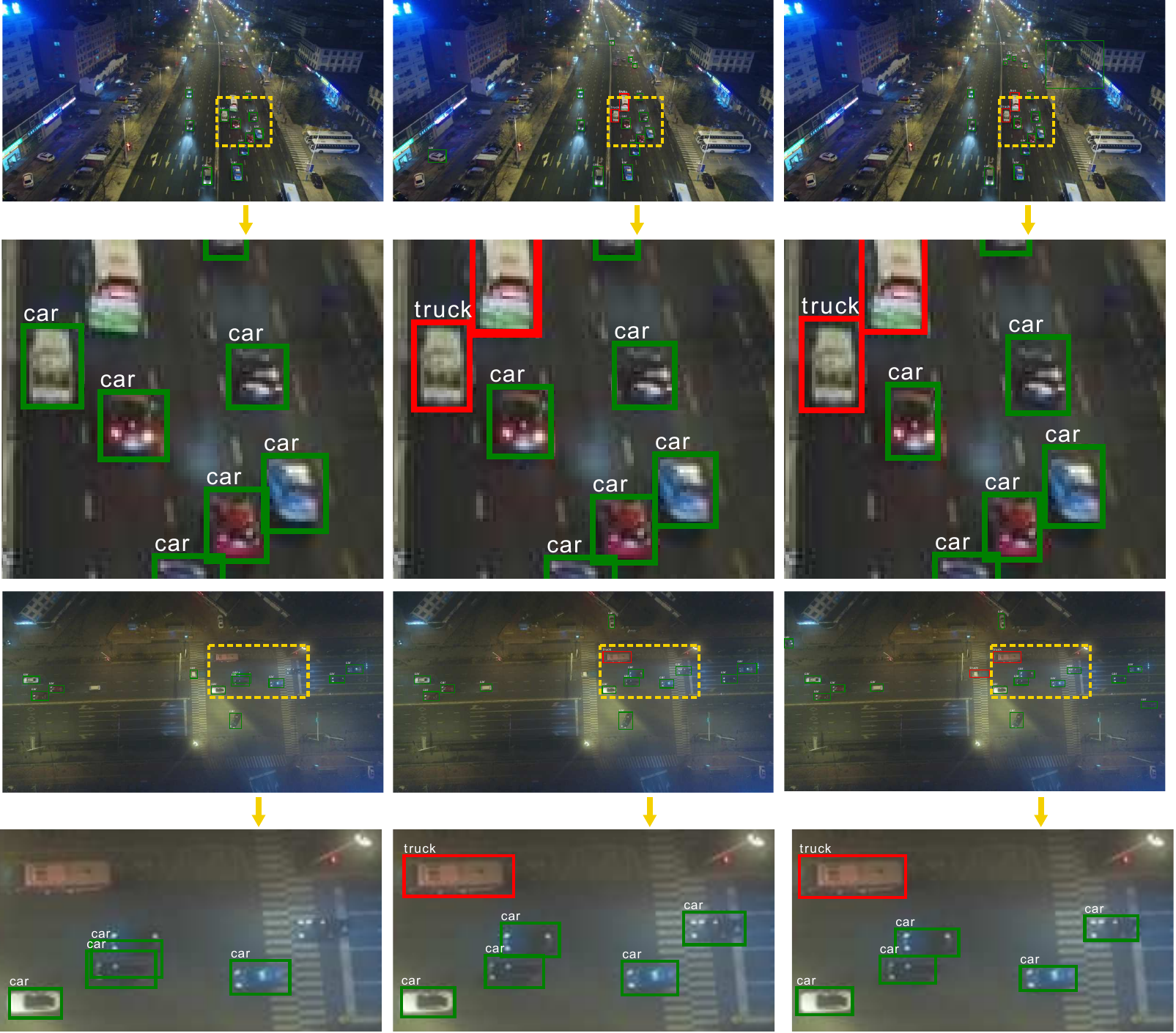}
\end{center}
\vspace{-0.1cm}
  \caption{Visualization of detection results on UAVDT. From the first to third columns are base model (Faster R-CNN (ResNet50)), DSHNet on the base model, and ground truth (\textcolor{red}{red} boxes for tail classes and \textcolor{green}{green} boxes for head classes). Best viewed on screen with zoom.  
}
\label{fig:sup1}
\end{figure*}

\begin{figure*}
\begin{center}
 \includegraphics[width=0.98\linewidth]{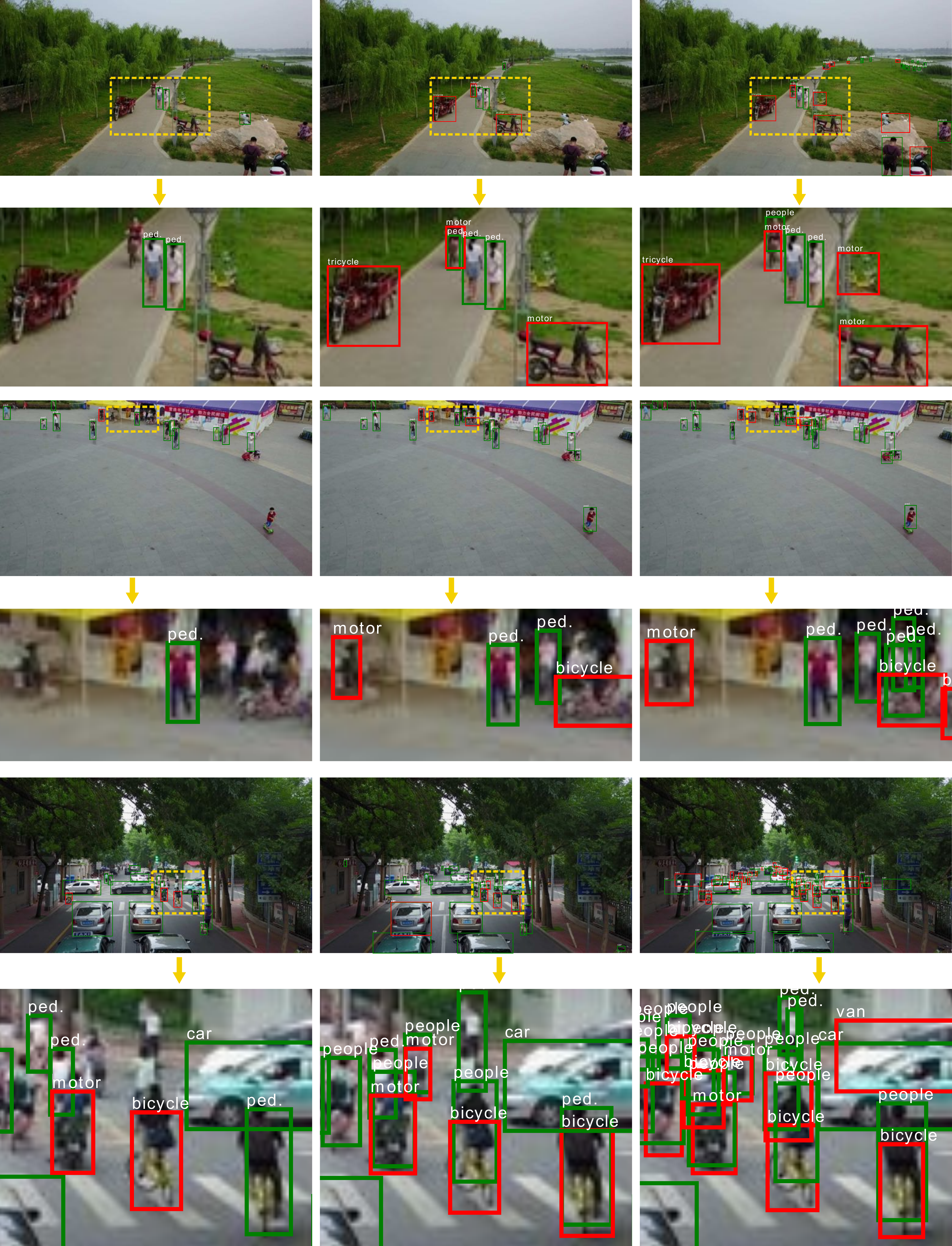}
\end{center}
\vspace{-0.1cm}
  \caption{Visualization of detection results on VisDrone. From the first to third columns are base model (Faster R-CNN (ResNet50)), DSHNet on the base model, and ground truth (\textcolor{red}{red} boxes for tail classes and \textcolor{green}{green} boxes for head classes). Best viewed on screen with zoom.  
}
\label{fig:sup2}
\end{figure*}

In this appendix, we provide more qualitative results of DSHNet compared with the base model.
Fig.~\ref{fig:sup1} and Fig.~\ref{fig:sup2} show the detection results on UAVDT and VisDrone, respectively.

\end{document}